\documentclass[runningheads]{llncs}

\usepackage{eccv}

\usepackage{eccvabbrv}

\usepackage{graphicx}
\usepackage{booktabs}
\usepackage{color}
\usepackage{bbding}

\usepackage[accsupp]{axessibility}  % Improves PDF readability for those with disabilities.

\usepackage{hyperref}

\usepackage{orcidlink}

\begin{document}

\title{Improving Text-guided Object Inpainting with Semantic Pre-inpainting} 

%\titlerunning{Text-guided Object Inpainting}

\author{Yifu Chen\inst{1,2}%\orcidlink{0009-0000-4272-3790} 
\and
Jingwen Chen\inst{3}%\orcidlink{0000-0002-7917-6003} 
\and
Yingwei Pan\inst{3}%\orcidlink{0000-0002-4344-8898} 
\and
Yehao Li\inst{3} \and
Ting Yao\inst{3} \and \\
Zhineng Chen\inst{1,2}$^{\dag}$ \and
Tao Mei\inst{3}%\orcidlink{0000-0002-5990-7307}
}

\authorrunning{Yifu Chen et al.}

\institute{School of Computer Science, Fudan University \and
 Shanghai Collaborative Innovation Center of Intelligent Visual Computing \and
HiDream.ai Inc. \\
\email{22210240129@m.fudan.edu.cn,~\{chenjingwen,~pandy,~liyehao,~tiyao\}@hidream.ai \\zhinchen@fudan.edu.cn, tmei@hidream.ai}}

\maketitle

\let\thefootnote\relax\footnotetext{$^{\ast}$ This work was performed when Yifu Chen was visiting HiDream.ai as a research intern. $^{\dag}$ Zhineng Chen is the corresponding author.}

\begin{abstract}
Recent years have witnessed the success of large text-to-image diffusion models and their remarkable potential to generate high-quality images. The further pursuit of enhancing the editability of images has sparked significant interest in the downstream task of inpainting a novel object described by a text prompt within a designated region in the image. Nevertheless, the problem is not trivial from two aspects: 1) Solely relying on one single U-Net to align text prompt and visual object across all the denoising timesteps is insufficient to generate desired objects; 2) The controllability of object generation is not guaranteed in the intricate sampling space of diffusion model. In this paper, we propose to decompose the typical single-stage object inpainting into two cascaded processes: 1) semantic pre-inpainting that infers the semantic features of desired objects in a multi-modal feature space; 2) high-fieldity object generation in diffusion latent space that pivots on such inpainted semantic features. To achieve this, we cascade a Transformer-based semantic inpainter and an object inpainting diffusion model, leading to a novel CAscaded Transformer-Diffusion (CAT-Diffusion) framework for text-guided object inpainting. Technically, the semantic inpainter is trained to predict the semantic features of the target object conditioning on unmasked context and text prompt. The outputs of the semantic inpainter then act as the informative visual prompts to guide high-fieldity object generation through a reference adapter layer, leading to controllable object inpainting. Extensive evaluations on OpenImages-V6 and MSCOCO validate the superiority of CAT-Diffusion against the state-of-the-art methods. Code is available at \url{https://github.com/Nnn-s/CATdiffusion}.

\end{abstract}

\section{Introduction}
\label{sec:intro}

\begin{figure}[t]
    \centering
    \includegraphics[width=0.85\linewidth]{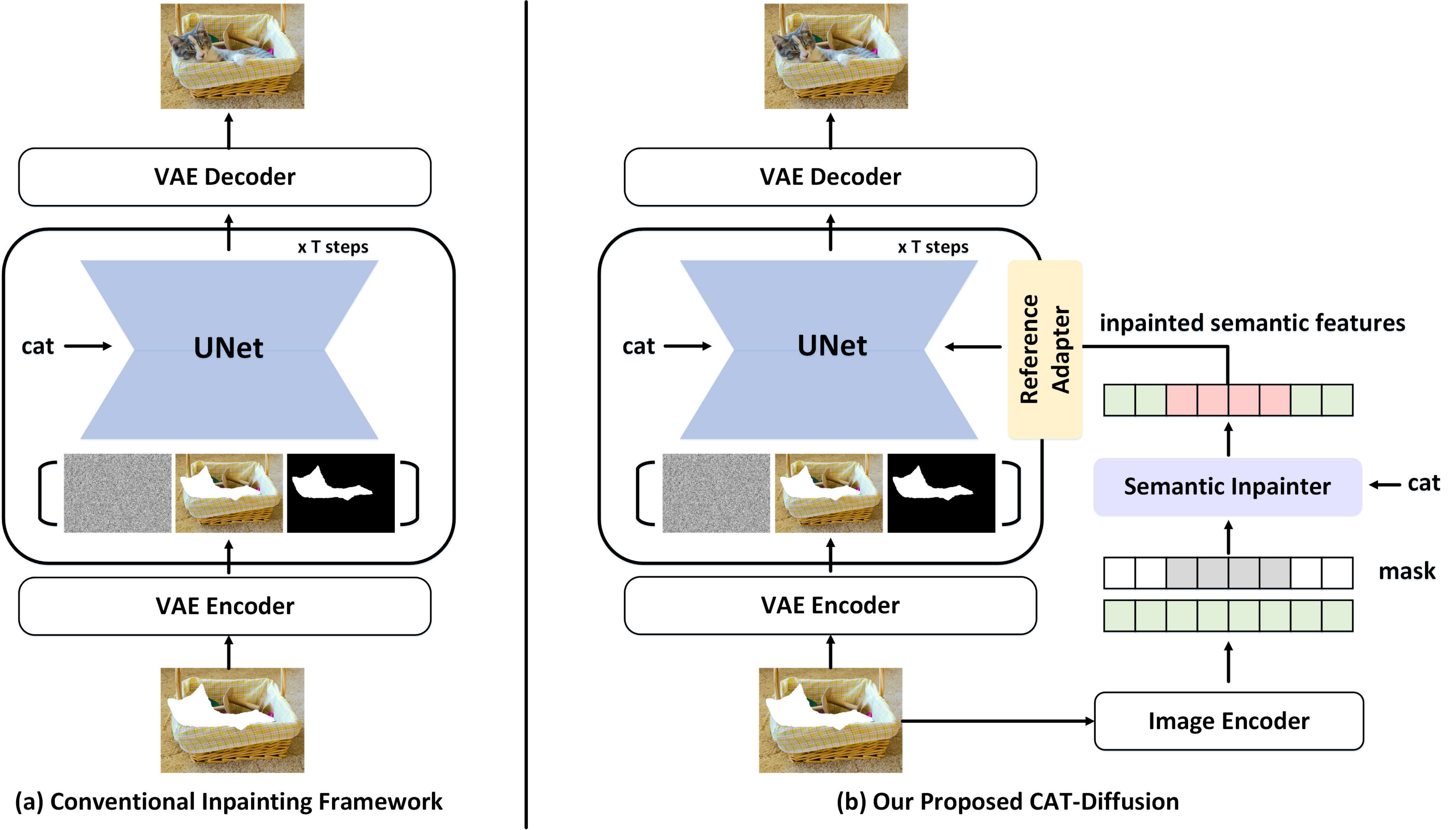}
    \caption{An illustration of conventional object inpainting framework and our proposed CAT-Diffusion. (a) Typical framework commonly feeds  a masked image with the original object removed, a text prompt describing the target object (e.g., ``cat'') and a binary mask indicating the designated region to be inpainted into an individual diffusion model for object inpainting. (b) Our proposed CAT-Diffusion additionally pre-inpaint the semantic features of the target object via Transformer-based semantic inpainter, and the derived features are leveraged to steer the diffusion model through the reference adapter layer for controllable and high-fidelity object inpainting.}
    \label{fig:intro}
\end{figure}

Traditional image inpainting \cite{ballester2001filling,bertalmio2000image,barnes2009patchmatch,criminisi2004region} focuses on restoring the corrupted area~of~an image without additional conditions, and is widely applied to object removal or damaged image repairing. Later works \cite{ZhangCHJ20textGuided,ZZHY20textGuided} have explored leveraging text prompts to complete the missing parts of an image, and shown impressive performances. Recent advances of diffusion-based text-to-image %synthesis 
models \cite{ho2020ddpm,rombach2022ldm,saharia2022photorealistic} have established a steady momentum to produce high-quality images conditioning on text prompts, and derived breakthroughs for image editing task of text-guided object inpainting \cite{nichol2021glide,avrahami2023blended,xie2023smartbrush}. Such inpainting task is to specifically restore a novel object within a masked region with the condition of a given text prompt.

One straightforward way for extending existing text-to-image diffusion models \cite{ho2020ddpm,rombach2022ldm} to text-guided object inpainting is to generate the new object from random noise and blend the object with background at every denoising step. Nevertheless, this method overlooks the inherent relations between new object and background, and might result in obvious artifacts around masked region and distorted object structure. The phenomenon motivates the subsequent efforts \cite{nichol2021glide,von-platen-etal-2022-diffusers} to finetune the text-to-image diffusion model via randomly~erasing part of the input image and enforcing the model to recover the missing area based on the associated caption. Despite having promising results, these attempts severely suffer from misalignment between the inpainting object and the given caption, due to the fact that the caption describes the holistic image rather than the local masked region. To address this issue, SmartBrush \cite{xie2023smartbrush} leverages the object label as prompt and the binary mask as the additional shape guidance to steer the diffusion model for object inpainting. The conventional framework of this kind of object inpainting diffusion models is illustrated in Figure \ref{fig:intro}(a).

In general, the difficulty of text-guided object inpainting in diffusion model originates from two perspectives: 1) Solely relying on a single U-Net without model ensembling \cite{feng2023ernie,xue2023raphael} is insufficient to generate text-aligned objects (i.e., align text prompt and image with varied noise ratios along diffusion process).
2) Precisely controlling the generation of visual objects without extra control signals remains challenging in the intricate sampling space of diffusion model. To mitigate the issues, we present a novel CAT-Diffusion to decompose the typical single-shot object inpainting into two cascaded stages: 1) first semantic pre-inpainting, i.e., inferring semantic feature of target object in a noise-free multi-modal feature space; 2) then object generation, i.e., leveraging the inferred semantic feature as visual prompt to steer diffusion model, thereby leading to high-fidelity object inpainting. Such design eases the learning of object-text alignment in diffusion model. Specifically, a Transformer-based semantic inpainter is first devised to predict the semantic features of the desired object conditioning on the unmasked image context and the text prompt. Note that an effective knowledge distillation objective in a multi-modal feature space (CLIP feature space) is leveraged to supervise the learning of semantic inpainter with ground-truth patch features. Furthermore, the outputs of the semantic inpainter, as the semantically-aligned and context-aware visual prompts, are fed into an object inpainting diffusion model through an additional reference adapter layer. This layer elegantly modulates the visual prompts to enhance the controllability of diffusion model for object inpainting. The overall framework of our proposed CAT-Diffusion is shown in Figure \ref{fig:intro}(b).

In summary, we have made the following contributions: 1) The proposed CAT-Diffusion is shown capable of high-fidelity object inpainting through two decomposed stages, i.e., first semantic pre-inpainting and then obejct generation; 2) The designed diffusion model equipped with a reference adapter layer is shown able to nicely control object inpainting by modulating the visual prompts (i.e., the inpainted semantic features); 3) CAT-Diffusion has been properly verified through extensive experiments over OpenImages-V6 and MSCOCO.

\section{Related Work}

\textbf{Traditional Image Inpainting.}
Conventional image inpainting task aims to synthesize visually coherent and plausible content within the corrupted region of an image. Traditional methods \cite{ballester2001filling,bertalmio2000image,barnes2009patchmatch,criminisi2004region} exploits the spatial redundancy of images to patch holes with image fragments sourced from elsewhere in the image, where the high-level semantics of the image are overlooked. To tackle this problem, recent inpainting techniques \cite{ZengFCG19pyd,QuanZZLWY22tip,YuLYSLH19gatedConv,LiuRSWTC18partialConv,Yu0YSLH18contextAttn} leverage the semantic prior learned from large-scale image data and dynamically model the global-local interactions in the image with deep models (e.g., Generative Adversarial Networks \cite{goodfellow2020gan,pan2017create}, Transformer-based architectures \cite{vaswani2017transformer,li2022contextual,yao2023dual,yao2024hiri}). However, the learning process will become increasingly challenging with higher image resolution. Thus, some works \cite{SongYLLHLK18divide,YiTAJX20residual,YangLLSWL17multi,LiLZQWJ22mat} are proposed to  resolve high-resolution image inpainting and promising results are achieved. Variational Auto-Encoder (VAE) \cite{kingma2013vae} based methods \cite{Peng0XL21vqvae,zhao2020uctgan,zheng2019pluralistic,LiuWHSH021diverse} further improve the diversity of outputs by sampling diverse and high-quality structural information from a latent space, and incorporate the derived structure encoding into the image generator for hole filling.

\noindent\textbf{Text-Guided Image Inpainting.}
Despite their impressive performances, the aforementioned methods rely on the assumption that sufficient contextual information will be available within the source image. However, this assumption may not always hold. To minimize the loss of semantic information and facilitate greater control over the inpainting process, text prompts are employed as extra conditions in \cite{ZhangCHJ20textGuided,ZZHY20textGuided}, thereby introducing a new task of text-guided object inpainting. This newly introduced task differs fundamentally from conventional image inpainting in requiring visual and semantic consistency with the source image and the text prompt, respectively. The emergence of diffusion models \cite{ho2020ddpm,rombach2022ldm,zhu2024sd} has propelled advancements in text-driven content creation with their powerful capability of high-fidelity image synthesis. Building upon this technology, recent studies \cite{xie2023smartbrush,avrahami2022blended,nichol2021glide,avrahami2023blended} showcase the ability to generate a novel object described by a text prompt within a region indicated by a binary mask. Furthermore, SmartBrush \cite{xie2023smartbrush} uses the object mask as additional guidance during the inpainting process to boost the performances.

\noindent\textbf{Diffusion Models.}
Diffusion denoising probabilistic models (DDPM) \cite{ho2020ddpm} have reshaped the landscape of image synthesis. However, notable challenges such as high computational demands and low inference efficiency arise, and hinder the deployment of DDPMs in practices. Significant efforts \cite{dhariwal2021diffusion,ho2022classifier,song2020denoising} have been dedicated to improve DDPMs and further tap their potentials. Particularly, Latent Diffusion Models (LDMs) \cite{rombach2022ldm} train the DDPM in a low-dimensional latent space learned by Variational Auto-Encoder, resulting in faster training and inference while producing high-fidelity images. With these noteworthy improvements, LDMs have solidified the position in the realms of text-to-image synthesis \cite{brooks2023instructpix2pix,hertz2022prompt,saharia2022photorealistic,zhang2023controlnet,chen2023controlstyle}, video generation \cite{ho2022imagen,wu2023tune,zhang2024trip}, 3D generation \cite{poole2022dreamfusion,tang2023make,chen2023control3d}.

\noindent\textbf{Summary.}
%{\color{blue}
In this paper, we investigate the problem of text-guided object inpainting and propose a novel Cascaded Transformer-Diffusion (CAT-Diffusion) framework. In contrast to conventional methods that leaning on an individual diffusion model, our CAT-Diffusion decomposes the single-shot pipeline into two cascaded stages: first semantic pre-inpainting and then object generation. Technically, the semantic features of the target object are pre-inpainted in a noise-free multi-modal feature space (CLIP space). Different from existing methods \cite{nichol2021glide,avrahami2022blended} that optimizes the latent code to achieve higher CLIP score between regional/global image and object prompt during inference, our CAT-Diffusion predicts the semantic features of the desired object that naturally align with the prompt by exploiting the mask-unmask visual context and pre-trained multi-modal priors in CLIP. Then, such inpainted semantic features are utilized to steer the diffusion model for controllable object generation, which eases the learning of object-text alignment in diffusion model across varied noise ratios.

\section{Approach}
\label{sec:tech}

To begin with, it is worth recalling that, in text-guided object inpainting, a novel object described by a text prompt (usually the object label) needs to be generated within a designated region indicated by a binary mask of an input image. Both visual coherence and semantic relevance with the image and the text prompt, respectively, are required in this task. In this section, we will delve into the specifics of our proposed Cascaded Transformer-Diffusion (CAT-Diffusion) after a brief review of diffusion models, followed by the training details.

\subsection{Prelimiaries}
\label{sec:ddpm_bg}

Denoising Diffusion Probabilistic Model (DDPM) \cite{ho2020ddpm}, a new type of generative model, can be regarded as a parameterized Markov chain able to produce images that closely match a desired data distribution within finite timesteps $T$. Generally, a forward diffusion process and a reverse denoising process are involved in DDPM. Specifically, in the forward diffusion process, DDPM gradually applies noise to the image according to a pre-defined variance schedule $\{\beta_t\}_1^T$ until the image is fully destroyed. Conversely, during the reverse denoising process, DDPM is trained to estimate the added noise to the noisy image and progressively remove it over a sequence of $T$ timesteps. The training objective of DDPM (usually implemented as a U-Net \cite{ronneberger2015unet}) parameterized by $\theta$ can be simply formulated as
\begin{equation}
    \label{eq:ddpm}
    \mathcal{L}_{ddpm}(\theta, x) = \mathbb{E}_{t \sim \mathcal{U}(0, T), \epsilon \sim \mathcal{N}(0,I)}[\left\|\epsilon_\theta(x_t, t, c) - \epsilon\right\|_2^2],
\end{equation}
where $\epsilon$ is a random noise sampled from the normal distribution $\mathcal{N}(0,I)$, $c$ is some kind of condition (e.g., text prompt), and $x_t$ is the noisy image at timestep $t$ derived as follows:
\begin{equation}
    x_t = \sqrt{\bar{\alpha_t}}x_0 + \sqrt{1 - \bar{\alpha_t}}\epsilon,
\end{equation}
where $\alpha_t = 1 - \beta_t$, $\bar{\alpha_t} = \prod_{s=1}^{t}\alpha_s$. To address the challenge of training DDPM at higher resolutions with limited GPU resources, Latent Diffusion Model (LDM) \cite{rombach2022ldm} instead trains DDPM in latent space of lower dimensions than pixel space. In this paper, we follow the similar recipe, and the training objective Eq. \ref{eq:ddpm} can be rewritten as:
\begin{equation}
    \label{eq:ldm}
    \mathcal{L}_{ldm}(\theta, z) = \mathbb{E}_{t \sim \mathcal{U}(0, T), \epsilon \sim \mathcal{N}(0,I)}[\left\|\epsilon_\theta(z_t, t, c) - \epsilon\right\|_2^2],
\end{equation}
where $z$ is the latent representation of $x$ and usually obtained from a pre-trained Variational Auto-Encoder (VAE).
%commonly implemented as a U-Net \cite{ronneberger2015unet},

\begin{figure}[t]
    \centering
    \includegraphics[width=0.96\linewidth]{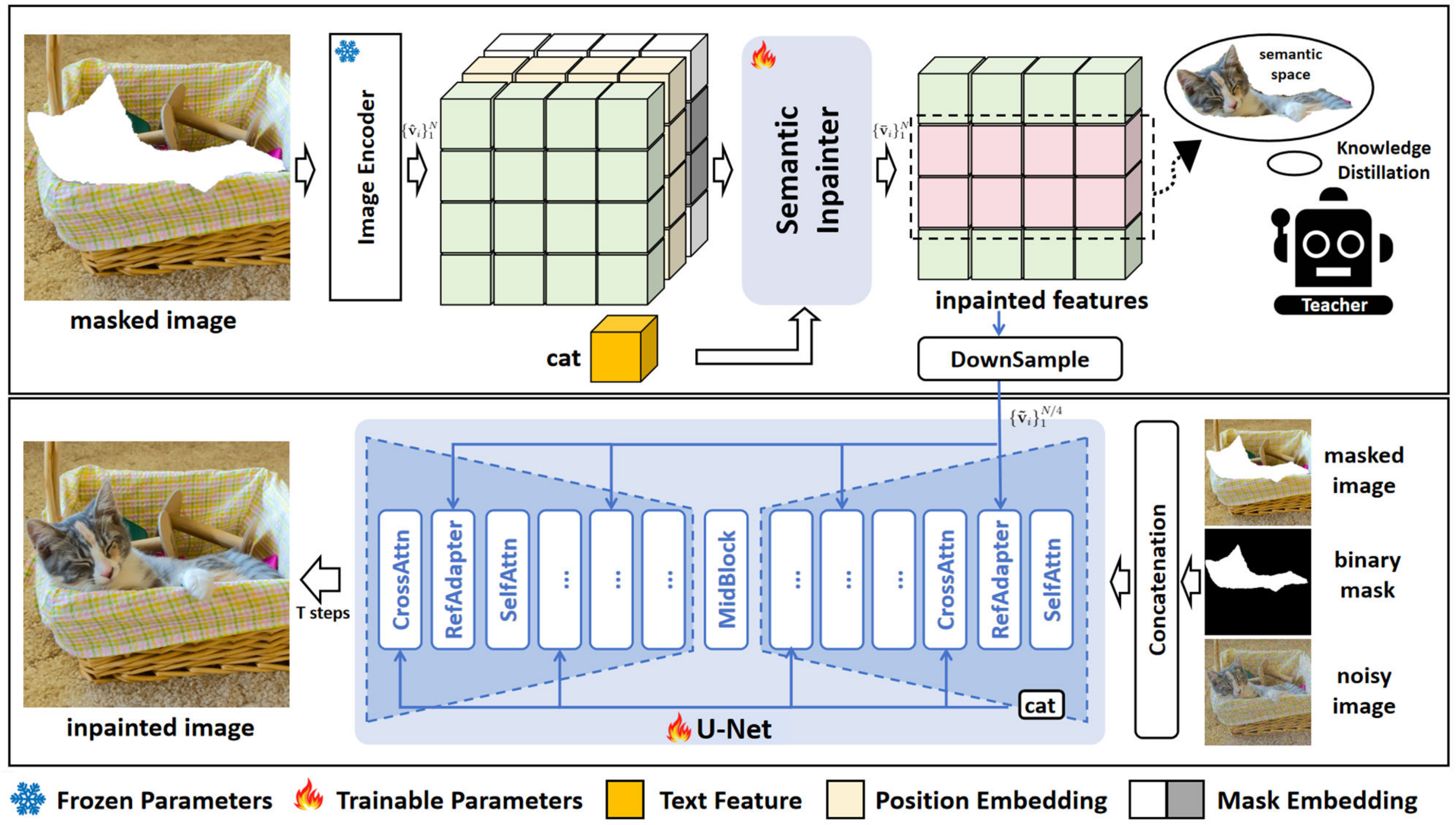}
    \caption{The framework of our CAT-Diffusion. Specifically, a pre-trained image encoder is first employed to extract the visual features of the masked image. Then, a novel semantic inpainter takes these visual features and a text prompt as inputs, and pre-inpaints the semantic features of the desired object in a multi-modal feature space, thereby aligning the prompt and the visual object in addition to the U-Net regardless of denoising timesteps. To achieve this goal, knowledge distillation is adopted to transfer the multi-modal knowledge from a teacher model to the semantic inpainter. Finally, an object inpainting diffusion model equipped with a reference adapter layer is steered by the aligned semantic features for controllable object inpainting in visual space.}
    \label{fig:framework}
\end{figure}

\subsection{Cascaded Transformer-Diffusion Model}
\label{sec:cat}
Let $x$, $c$ and $m$ be an input image, a text prompt that describes the novel object, and a binary mask that indicates the region to be inpainted, respectively. Most of the existing text-guided object inpainting diffusion models \cite{avrahami2023blended,nichol2021glide} are merely optimized to align the text prompt $c$ and the desired object $x \odot m$ in the latent space, which leads to suboptimal results. As discussed previously in Section. \ref{sec:intro}, two primary issues need to be addressed for further improvements: 1) Relying on a standalone U-Net to achieve visual-semantic alignment across all denoising timesteps is insufficient \cite{feng2023ernie,xue2023raphael} for text-guided object inpainting; 2) It is challenging to stably generate high-fidelity objects in the intricate sampling space without extra semantic information.

%{\color{blue}
To overcome these challenges, we propose to decompose the conventional single-stage pipeline into two cascaded stages: first semantic pre-inpainting and then object generation, leading to our CAT-Diffusion. Technically, our CAT-Diffusion pre-inpaints the object in an auxiliary multi-modal feature space (e.g., CLIP \cite{radford2021clip}) via a novel semantic inpainter. The semantic inpainter is trained through knowledge distillation \cite{navaneet2022simreg} to predict the semantic features of the target object conditioning on the unmasked visual context and text prompt. This way, the derived outputs naturally align the text prompt and the visual object in addition to the U-Net regardless of the denoising timesteps. The outputs from the semantic inpainter is further integrated into an object inpainting diffusion model through a reference adapter layer for controllable object inpainting. The overall framework of CAT-Diffusion is shown in Figure \ref{fig:framework}.
%}

\subsubsection{Semantic Inpainter.}

To alleviate the deficiency in aligning text prompt and visual object solely depending on a standalone U-Net across the entire denoising process, we propose to enhance the visual-semantic correspondence in a well pre-trained auxiliary multi-modal feature space in addition to the U-Net by pre-inpainting the semantic features of the desired object. The rationale behind is that the pre-trained multi-modal feature space is learned on large-scale cross-modal data for visual-semantic alignment regardless of the denoising timesteps. In this work, an effective knowledge distillation objective is devised to transfer this kind of multi-modal knowledge from a teacher model (CLIP) to the semantic inpainter in our CAT-Diffusion.

Specifically, a Transformer-based semantic inpainter ($SemInpainter$) parameterized by $\eta$ is cascaded with a object inpainting diffusion model in our CAT-Diffusion, as shown in the upper part of Figure \ref{fig:framework}. First, the visual features $\{\mathbf{\hat{v}}_i\}_1^N$ of the masked image $x \odot (1 - m)$ are extracted by the image encoder of CLIP ($ImageEnc$), where $\mathbf{\hat{v}}_i \in \mathbb{R}^d$ is a feature vector of a local image patch. Given the corrupted features $\{\mathbf{\hat{v}}_i\}_1^N$ of the masked image and the textual feature $\mathbf{c} \in \mathbb{R}^d$ of the text prompt $c$, $SemInpainter$ aims to predict semantic features $\{\mathbf{\bar{v}}_i\}_1^N$ that faithfully reconstruct the target object in CLIP space. This way, the visual-semantic alignment is naturally attained along the entire denoising process of diffusion model. This whole process can be expressed as
\begin{equation} 
    \begin{aligned}
        \{\mathbf{\bar{v}}_i\}_1^N & = SemInpainter([\{\mathbf{\hat{v}}_i\}_1^N + \mathbf{PE} + \mathbf{ME}, \mathbf{c}] ), \\
        \{\mathbf{\hat{v}}_i\}_1^N & = ImageEnc(x \odot (1 - m)),
    \end{aligned}
\end{equation}
where $[~\cdot~]$ denotes the concatenation operation, $\mathbf{PE} \in \mathbb{R}^{N \times d}$ is a learnable position embedding, and $\mathbf{ME} \in \mathbb{R}^{N \times d}$ is a learnable mask embedding that indicates whether the visual feature $\mathbf{\hat{v}}_i$ is masked in the image $x \odot (1 - m)$ or not. Our $SemInpainter$ is implemented as 24-layer Transformer structure (similar to the CLIP \cite{radford2021clip} image encoder). Finally, we derive the inpainted features $\{\mathbf{\tilde{v}}_i\}_1^N$, which will be later incorporated into the object inpainting diffusion model, by blending $\{\mathbf{\hat{v}}_i\}_1^N$ and $\{\mathbf{\bar{v}}_i\}_1^N$ with the binary mask $m$:
\begin{equation}
    \label{eq:blending}
    \{\mathbf{\tilde{v}}_i\}_1^{N/4} = DownSample(\{\mathbf{\bar{v}}_i\}_1^N \odot m + \{\mathbf{\hat{v}}_i\}_1^N \odot (1 - m)),
\end{equation}
where $DownSample(~\cdot~)$ is a parameter-free downsample layer to reduce the feature resolution for computation efficiency.

To enforce the predicted object features aligning with the text prompt $c$, we propose to transfer the multi-modal knowledge from a teacher model (CLIP is adopted as the teacher model in this paper) to the $SemInpainter$. Technically, the $SemInpainter$ is trained with the task of Masked Feature Prediction \cite{navaneet2022simreg,wei2022masked} to recover the ground-truth semantic features of the masked object in CLIP space conditioned on the unmasked visual context and the text prompt $c$.

\subsubsection{Reference Adapter Layer.}
Since controlling the diffusion model for high-fidelity text-to-image generation remains challenging, inspired by \cite{zhang2023controlnet,zhao2023unicontrol}, the semantic features $\{\mathbf{\tilde{v}}_i\}_1^{N/4}$ are utilized as additional conditions, referred as visual prompts, to steer the diffusion model for controllable inpainting in latent space.

Specifically, we introduce a new reference adapter layer ($RefAdapter$) to the base object inpainting diffusion model, which is formulated as a multi-head cross-attention layer of Transformer style. The proposed $RefAdapter$ is alternately plugged in between the original self-attention ($SelfAttn$) and cross-attention layers ($CrossAttn$) of the U-Net, as shown in the lower part of Figure \ref{fig:framework}. Let $\mathbf{H} \in \mathbb{R}^{N_h \times d_h}$ be the intermediate hidden states of the U-Net, where $N_h$ vary along with the resolution of different layers. $\mathbf{C} \in \mathbb{R}^{N_c \times d_c}$ is the textual features of the prompt $c$ extracted by the text encoder in the diffusion model. A shared module prototype for different types of attention layers is first defined as:
\begin{equation}
    Output = Module(Query, Key, Value).
\end{equation}
Therefore, a full block in the upgraded U-Net including $SelfAttn$, $RefAdapter$ and $CrossAttn$ operates as
\begin{equation}
    \begin{aligned}
        \mathbf{H}_x & = CrossAttn(\mathbf{H}_r, \mathbf{C}, \mathbf{C}) + \mathbf{H}_r,                                              \\
        \mathbf{H}_r & = RefAdapter(\mathbf{H}_s, \{\mathbf{\tilde{v}}_i\}_1^{N/4}, \{\mathbf{\tilde{v}}_i\}_1^{N/4}) + \mathbf{H}_s, \\
        \mathbf{H}_s & = SelfAttn(\mathbf{H}, \mathbf{H}, \mathbf{H}) + \mathbf{H},
    \end{aligned}
\end{equation}
where $\mathbf{H}_x$ will be the input to the next block. 

\subsubsection{Object Inpainting Diffusion Model.}
By decomposing the object inpainting pipeline into the two cascaded processes, our CAT-Diffusion can generate high-fidelity objects that are semantically aligned with the text prompt. Moreover, the visual coherence between the filled-in object and the surrounding image context is further enhanced by steering the diffusion model with additional semantic guidance (i.e., the inpainted visual features $\{\mathbf{\tilde{v}}_i\}_1^{N/4}$).

Generally, an inpainting diffusion model with U-Net parameterized as $\epsilon_\theta$ takes the noisy latent code $z_t$, the mask $m$ indicating the region to be inpainted and the masked latent code $z \odot (1 - m)$ as inputs, and estimates the noise to be removed from $z_t$, leading to the denoised one $z_{t-1}$. Mathematically, the overall procedure can be expressed as
\begin{equation}
    %\begin{aligned}
        z_{t-1} = \frac{1}{\sqrt{\alpha_{t}}}(z_t - \frac{1 - \alpha_t}{\sqrt{1 - \bar{\alpha_t}}} \cdot \epsilon_\theta([z_t, z \odot (1 - m),m], t, c,\{\mathbf{\tilde{v}}_i\}_1^{N/4})) + \sigma_t \epsilon.
    %\end{aligned}
\end{equation}
The readers can refer to \cite{ho2020ddpm,rombach2022ldm} for more details.

\subsection{Training}
\label{sec:training}

{\textbf{Diffusion Loss.}}
For the training of the object inpainting diffusion model equipped with the reference adapter layer, we employ the general practice in \cite{rombach2022ldm} following the objective function as
\begin{equation}
    \label{eq:train_ldm}
    %\begin{aligned}
        \mathcal{L}_{ldm} (\theta, z, m) = \mathbb{E}_{t \sim \mathcal{U}(0, T), \epsilon \sim \mathcal{N}(0,I)} [\|\epsilon_\theta([z_t, z \odot (1 - m), m], t, c, \{\mathbf{\tilde{v}}_i\}_1^{N/4}) - \epsilon\|_2^2].
    %\end{aligned}
\end{equation}

\noindent{\textbf{Knowledge Distillation Loss.}}
\iffalse
Moreover, to enforce the semantic inpainter to pre-inpaint the object features that are semantically aligned with the prompt $c$ and coherent to the visual context, we surpervise the pre-training and finetuning of the semantic inpainter with the ground-truth visual features $CLIP(x)$ as in the task of Masked Feature Prediction \cite{navaneet2022simreg,wei2022masked}. The optimization objective can be formulated as
\begin{equation}
    \begin{aligned}
    \mathcal{L}_{distill}(\eta, x, m) = \mathbb{E}_{(x, c) \sim \{\mathcal{X},\mathcal{C}\}} [\| SemInpainter_{\eta}(ImageEnc&(x \odot (1 - m)), c) \\
& - CLIP(x)  \|_2^2 ],
    \end{aligned}
\end{equation}
where $\{\mathcal{X},\mathcal{C}\}$ is the training dataset of paired image $x$ and text prompt $c$, and $\eta$ denotes the trainable parameters of the proposed $SemInpainter$ in Section. \ref{sec:cat}.
\fi
Moreover, to enforce the semantic inpainter to pre-inpaint the object features that are semantically aligned with the prompt $c$, we supervise the semantic inpainter with the ground-truth visual features $CLIP(x)$ in the task of Masked Feature Prediction. The optimization objective can be formulated as
\begin{equation}
    \begin{aligned}
    \mathcal{L}_{distill}(\eta, x, m) = \mathbb{E}_{(x, c) \sim \{\mathcal{X},\mathcal{C}\}} [\| SemInpainter_{\eta}(ImageEnc&(x \odot (1 - m)), c) \\
& - CLIP(x)  \|_2^2 ],
    \end{aligned}
\end{equation}
where $\{\mathcal{X},\mathcal{C}\}$ is the training dataset of paired image $x$ and text prompt $c$, and $\eta$ denotes the trainable parameters of the proposed $SemInpainter$ in Section. \ref{sec:cat}.

\section{Experiments}

We verify the merit of the proposed Cascaded Transformer-Diffusion approach (CAT-Diffusion) for the task of text-guided object inpainting and compare with state-of-the-art diffusion-based methods \cite{avrahami2022blended,nichol2021glide,avrahami2023blended,rombach2022ldm,xie2023smartbrush}. Extensive experiments validate the effectiveness of CAT-Diffusion in inpainting high-fidelity objects.

\subsection{Implementation Details}
Following \cite{xie2023smartbrush}, we train CAT-Diffusion on the pairs of local mask and the corresponding object label from the training set of OpenImages-V6 \cite{kuznetsova2020openimages}. CAT-Diffusion is optimized by Adam \cite{kingma2014adam} with learning rate 0.00001 for about 40K iterations on 8 A100 GPUs. Batch size is set to 128 and the input image resolution is set as 512 $\times$ 512.

\subsection{Compared Methods and Evaluation Metrics}
\subsubsection{Compared Methods.}
We compare CAT-Diffusion with several state-of-the-art diffusion-based approaches including \textbf{Blended Diffusion} \cite{avrahami2022blended}, \textbf{Blended Latent Diffusion} \cite{avrahami2023blended}, \textbf{GLIDE} \cite{nichol2021glide}, \textbf{SmartBrush} \cite{xie2023smartbrush}, \textbf{Stable Diffusion} \cite{rombach2022ldm} and \textbf{Stable Diffusion Inpainting}. Specifically, Blended Diffusion, Blended Latent Diffusion and Stable Diffusion simply leverage a pre-trained base text-to-image model for text-guided object inpainting by blending the generated object and the background at each denoising step. The other methods train an inpainting diffusion model with the text prompt, binary mask and masked image as the inputs. Due to the same evaluation setup, the results of all methods are taken from \cite{xie2023smartbrush}, except for \cite{avrahami2023blended}. Please note that we have replaced the text-to-image stable diffusion 2.1 with 1.5 in Blended Latent Diffusion for fair comparisons. 

\subsubsection{Evaluation Metrics.}
All the aforementioned methods are evaluated on the test sets of both OpenImages-V6 \cite{kuznetsova2020openimages} and MSCOCO \cite{lin2014mscoco}, leading to 13,400 and 9,311 test images, respectively. We adopt three widely-used metrics: Frechet Inception Distance (FID) \cite{HeuselRUNH17FID,Seitzer2020FID}, Local FID and CLIP score \cite{hessel2021clipscore}. It is worth mentioning that FID and Local FID measure the realism and visual coherence of the inpainted object in global image and local patch respectively, while CLIP score estimates the semantic relevance between the inpainted object and the text prompt. User study is additionally involved to evaluate both the visual coherence and text-object alignment. Since GLIDE only supports images of resolution 256 $\times$ 256, we resize all the results to the similar size for fair comparisons. Moreover, both segmentation mask and bounding box mask are considered in evaluation.

\subsection{Performance Comparisons}

\subsubsection{Quantitative Results on OpenImages-V6.}
Table \ref{tab:openimages} summarizes the results of all the methods on test set of OpenImages-V6. Overall, results across all the metrics consistently demonstrate the effectiveness of our proposed CAT-Diffusion with segmentation mask or bounding box mask. Specifically, blending-based approaches (i.e., Blended Latent Diffusion and Stable Diffusion) achieve comparable CLIP scores but much inferior FID and Local FID scores to our CAT-Diffusion. We speculate that this is because these methods merely focus on the visual-semantic alignment between the inpainted image and the object label, and simply blend the generated object and background in the latent space. As such, the semantic context of the surrounding unmasked regions is overlooked, resulting in poor visual coherence. SmartBrush, by incorporating the masked image into U-Net for context learning and further guiding the diffusion model with shape mask, exhibits better performances. However, the FID and Local FID scores of SmartBrush are still below those of our CAT-Diffusion. The results verify the impact of steering the diffusion model with the pre-inpainted object features from the semantic inpainter through the reference adapter layer.

\subsubsection{Quantitative Results on MSCOCO.}
Table \ref{tab:mscoco} lists the results of all the approaches on the test set of MSCOCO. It is noteworthy that neither SmartBrush nor our CAT-Diffusion are trained on MSCOCO. Similar to the trends on OpenImages-V6, CAT-Diffusion outperforms the other methods across most of the metrics. Specifically, CAT-Diffusion leads to the relative improvements over the strong baselines Stable Diffusion Inpainting and SmartBrush by 42.1\% and 20.7\% on Local FID (with bounding box mask), respectively. The results again validate the merit of decomposing the single-shot inpainting pipeline into two cascaded processes (first semantic pre-inpainting and then object generation) in our CAT-Diffusion.

\begin{table}[t]
    \centering
    \caption{Quantitative performance comparisons on the test set of OpenImages-V6.}
    \label{tab:openimages}
    \begin{tabular}{cccccccc}
        \toprule
        & \multicolumn{3}{c}{With Segmentation Mask} &   & \multicolumn{3}{c}{With Bounding Box Mask}                                                                                                   \\ \cline{2-4} \cline{6-8}
        Model     & FID $\downarrow$ & Local FID $\downarrow$ & CLIP $\uparrow$ &  & FID $\downarrow$ & Local FID $\downarrow$ & CLIP $\uparrow$  \\
        \midrule
        Blended Diffusion \cite{avrahami2022blended} & 9.72  & 21.93 & 0.261 &  & 11.05 & 29.16 & 0.265   \\
        GLIDE \cite{nichol2021glide}  & 9.03 & 21.09  & 0.250 &   & 9.70 & 22.45 & 0.252  \\
        Blended Latent Diffusion$\dagger$ \cite{avrahami2023blended} & 8.30 & 15.28 & 0.262  &  & 14.86 & 25.46 & 0.262  \\
        Stable Diffusion \cite{rombach2022ldm}  & 6.90  & 12.27 & \textbf{0.263}   &  & 9.10 & 15.28 & 0.265  \\
        Stable Inpainting \cite{rombach2022ldm} & 5.84 & 10.98 & 0.261 &  & 7.07 & 12.57  & 0.264  \\
        SmartBrush \cite{xie2023smartbrush}  & 4.70 & 7.82 & \textbf{0.263}  &  & 6.00 & 9.71 & \textbf{0.266}  \\
        \midrule
         CAT-Diffusion (ours) & \textbf{3.99}  & \textbf{6.32} & \textbf{0.263}  &  & \textbf{5.02} & \textbf{7.76}  & 0.265  \\
        \bottomrule
    \end{tabular}
\end{table}

\begin{table*}[t]
    \centering
    \caption{Quantitative performance comparisons on the test set of MSCOCO.}
    \label{tab:mscoco}
    \begin{tabular}{cccccccc}
        \toprule
        & \multicolumn{3}{c}{With Segmentation Mask} &  & \multicolumn{3}{c}{With Bounding Box Mask}\\ \cline{2-4} \cline{6-8}
        Model  & FID $\downarrow$ & Local FID $\downarrow$ & CLIP $\uparrow$  &  & FID $\downarrow$ & Local FID $\downarrow$ & CLIP $\uparrow$ \\
        \midrule
        Blended Diffusion \cite{avrahami2022blended}& 8.16  & 26.25  & 0.244  &   & 12.68 & 41.43 & 0.251 \\
        Blended Latent Diffusion$\dagger$ \cite{avrahami2023blended} & 7.88 & 19.84  & 0.246  &  & 12.40 & 29.53 & 0.248   \\
        Stable Diffusion \cite{rombach2022ldm} & 7.78 & 17.16 & 0.246   &  & 12.29 & 25.61 & 0.250  \\
        GLIDE \cite{nichol2021glide}  & 6.98 & 24.25  & 0.235   &  & 9.32  & 30.72  & 0.241   \\
        Stable Inpainting \cite{rombach2022ldm}  & 6.54 & 15.16  & 0.243   &  & 8.50  & 18.13   & 0.246  \\
        SmartBrush \cite{xie2023smartbrush}  & 5.76   & 9.80   & \textbf{0.249}   &  & 8.05  & 13.22   & \textbf{0.252}  \\
        \midrule
        CAT-Diffusion (ours)  & \textbf{5.35}  & \textbf{8.53}  & \textbf{0.249} &   & \textbf{6.84} & \textbf{10.49} & 0.251              \\
        \bottomrule
    \end{tabular}
\end{table*}

\subsubsection{Qualitative Comparisons.}
\begin{figure}[t]
    \centering
    \includegraphics[width=0.96\linewidth]{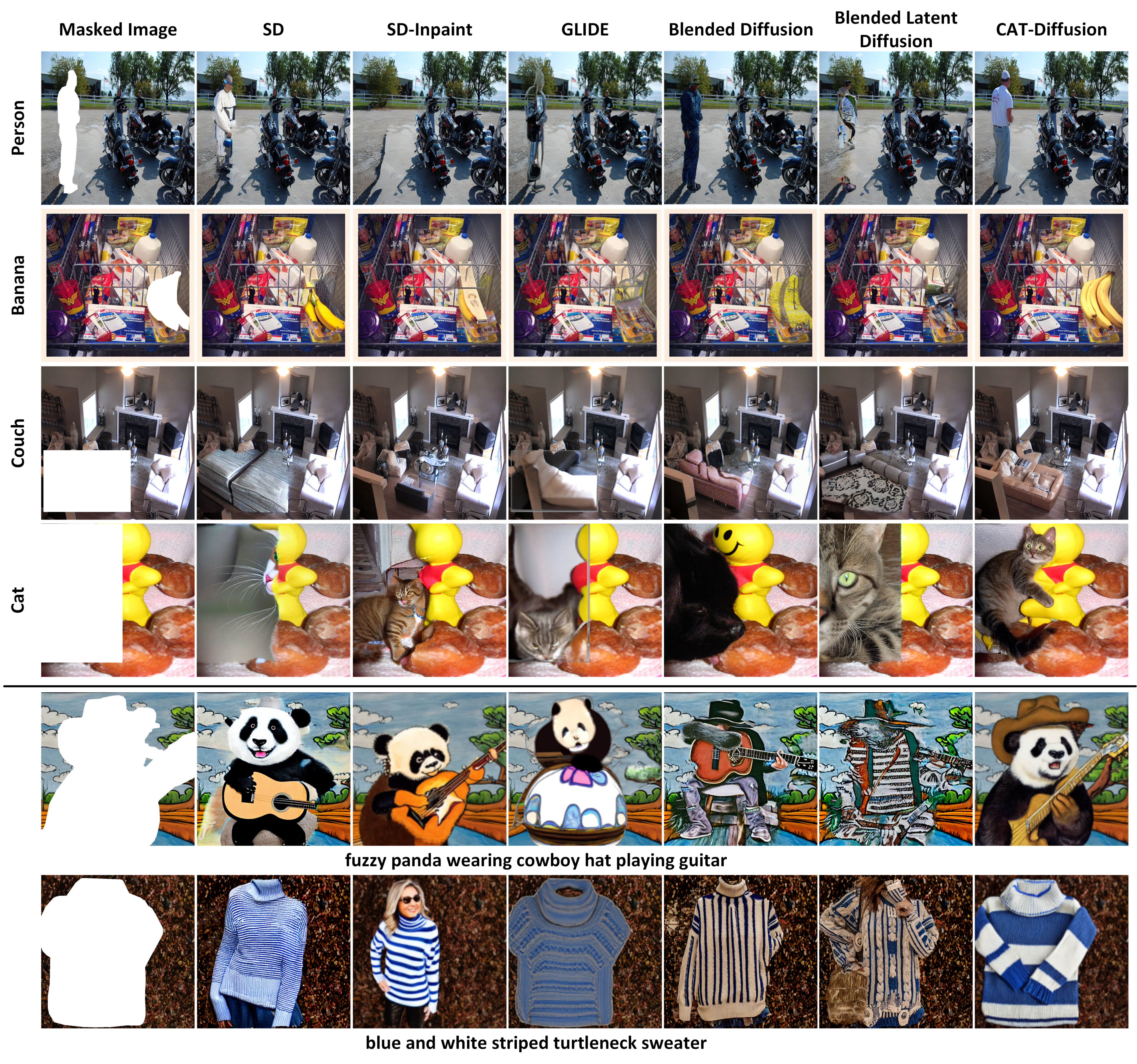}
    \caption{Examples generated by Stable Diffusion, Stable Diffusion Inpainting, GLIDE, Blended Diffusion, Blended Latent Diffusion and our proposed CAT-Diffusion with segmentation mask or bounding box mask.}
    \label{fig:examples}
\end{figure}

Then, we qualitatively test different methods through case study. Figure \ref{fig:examples} showcases several examples. As depicted by the top four results, the generated images by our CAT-Diffusion show better semantic alignment with the input text prompt than the other approaches. Moreover, better visual coherence of the objects with the surrounding context in the images and more accurate object shapes are also observed in the inpainted results of our CAT-Diffusion. The results prove the superiority of pre-inpainting the semantic features of the objects through the proposed semantic inpainter. For instance, compared to the images produced by other methods, the man generated by CAT-Diffusion at the first row is more structurally complete. This benefits from steering the diffusion model with the pre-inpainted semantic features of the target object through the reference adapter layer in our CAT-Diffusion. Though there is no shape mask provided, our CAT-Diffusion is able to generate high-fidelity objects given the text prompts and the bounding box masks (the middle two rows). Additionally, we evaluate object inpainting with more descriptive text prompts and the generated results of different methods are shown in the bottom two rows. Similarly, our CAT-Diffusion produces more visually pleasing images.

\begin{table}[t]
    \setlength{\tabcolsep}{5pt}
    \centering
    \caption{User study on 1K images randomly sampled from the test set of OpenImages-V6. BLD, SD and SD-Inapint are short for Blended Latent Diffusion, Stable Diffusion and Stable Diffusion Inpainting, respectively.}
    \label{tab:human}
    \begin{tabular}{cccccc}
        \toprule
                                     & \small GLIDE & \small BLD & \small SD & \small SD-Inpaint & \small CAT-Diffusion \\
        \midrule
        \small Text-Object Alignment & 1.1 & 1.1 &  1.0         &    2.2             &    3.5                  \\
        \small Visual Coherence       & 1.2 &  1.2 & 1.1           &   2.6               &  3.6                   \\
        \bottomrule
    \end{tabular}
\end{table}

\subsubsection{User Study.}
Next, we carry out a user study to examine whether the inpainting images conform to human preferences. In the experiments, we randomly sample 1K images from the test set of OpenImages-V6 for evaluation. Note that SmartBrush \cite{xie2023smartbrush} is not released and excluded here. We invite 10 evaluators (5 males and 5 females) with diverse education backgrounds: art design (4), psychology (2), computer science (2), and business (2). We show all the evaluators the inpainted images and the associated prompts, and ask them to assign scores (0 $\sim$ 5) from two aspects: 1) Visual coherence with the surrounding context; 2) Alignment with the text prompt and the accurate object shape. Table \ref{tab:human} summarizes the averaged results of different approaches. As indicated by the results, CAT-Diffusion leads the competition by a large margin against the other baselines in terms of both text-object alignment and visual coherence.

\subsection{Analysis and Discussions}
\subsubsection{Ablation Study on CAT-Diffusion.}
Here, we study how each component in CAT-Diffusion influences the overall performance. We consider one or more component at each stage and Table \ref{tab:ablation} summarizes the results on the test set of OpenImages-V6 with segmentation mask. Note that the baseline at Row \#1 is a Stable Inpainting model finetuned with object-text pairs from \cite{kuznetsova2020openimages}. By incorporating the reference adapter layer that is trained with CLIP features of masked images $x \odot (1 - m)$ only, the variant at Row \#2 makes the absolute improvement over the base model at Row \#1 by 0.91 and 1.47 on FID and Local FID scores, respectively. This is not surprising as the CLIP features of the unmasked region endow the base diffusion model with richer contextual semantics through the reference adapter layer, improving visual coherence and preserving the background. The outputs of semantic inpainter further boost the model with the semantics of the desired object, obtaining the best results at Row \#3 across all the metrics. 

\begin{table}[t]
   \setlength{\tabcolsep}{5pt}
    \centering
    \caption{Ablation study of the proposed CAT-Diffusion on the test set of OpenImages-V6 with segmentation mask.}
    \label{tab:ablation}
    \begin{tabular}{cccccc}
        \toprule
        \# & RefAd & SemInpt  & FID $\downarrow$ & Local FID $\downarrow$ & CLIP $\uparrow$ \\
        \midrule
        1  &                  &               & 5.15                    & 8.80    &  0.262                 \\
        2  & \Checkmark       &               & 4.24                    & 7.33    &  0.262                 \\
        3  & \Checkmark       & \Checkmark     & 3.99                    & 6.32   &  0.263                 \\
        \bottomrule
    \end{tabular}
\end{table}

\begin{figure}[t]
    \centering
    \includegraphics[width=0.9\linewidth]{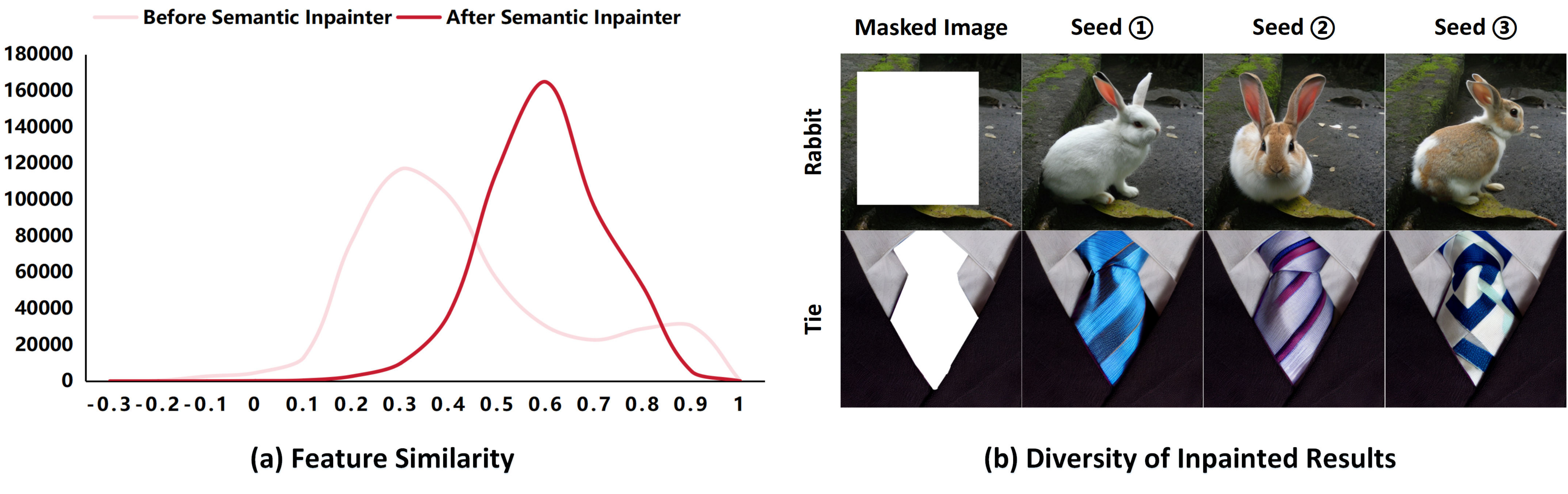}
    \caption{(a) The distributions of cosine similarity between the semantic features before/after the semantic inapinter within the masked region and the corresponding ground-truth ones. (b) Diverse inpainted images generated by our proposed CAT-Diffusion using different random seeds.}
    \label{fig:analysis}
\end{figure}

\subsubsection{Predicted Features by Semantic Inpainter.}
We then analyze the extent to which the proposed semantic inpainter improves the semantic features of the desired objects. It is worth noting that the CLIP features of the masked region inherently encode the contextual semantics from the unmasked area due to self-attention mechanism in CLIP, resulting in non-trivial similarity with the ground-truth ones before passing through semantic inpainter. In particular, we compute the cosine similarity between the inputs/outputs of the semantic inpainter within the masked region and the corresponding ground-truth ones on 10K images. Figure \ref{fig:analysis}(a) depicts the two distributions. The averaged cosine similarity is boosted from 0.47 to 0.65, demonstrating the effectiveness of the proposed semantic inpainter. Though the outputs from semantic inpainter are not 100\% accurate, richer context is contributed by these semantic features to generate high-fidelity objects in CAT-Diffusion.  

\subsubsection{Diversity of Inpainted Results.}
In order to test the diversity of the inpainted results by our CAT-Diffusion with the same semantic features derived from the semantic inpainter, we conduct a study on the results with different random seeds. Figure \ref{fig:analysis}(b) shows two examples. It can be observed that our CAT-Diffusion is able to inpaint diverse objects with accurate shapes controlled by the reference adapter layer.

%{\color{blue}
\subsubsection{Inference Complexity.}
During inference, we only require one-pass forward of the proposed semantic inpainter and the inpainted features can be reused in each denoising step, leading to minor computation overhead. The average time of our CAT-Diffusion (1.84 secs) for each image compared with SD-Inpaint (1.60 secs).

%\noindent\textbf{Limitations}
%XXX
%}

\section{Conclusion}
In this paper, a novel Cascaded Transformer-Diffusion (CAT-Diffusion) model is proposed to enhance the visual-semantic alignment and controllability of diffusion model for text-guided object inpainting. Specifically, our CAT-Diffusion decomposes the conventional single-stage pipeline into two cascaded processes: first semantic pre-inpainting and then object generation. By pre-inpainting the semantic features of the disired objects in a multi-modal feature space and then guiding the diffusion model with these features for object generation, our CAT-Diffusion is able to generate high-fidelity objects semantically aligned with the prompt and visually coherent with the background. Technically, a Transformer-based semantic inpainter predicts the semantic features of the desired object given the unmasked context and the prompt. Then, the inpainted features from the semantic inpainter are further fed into the object inpainting diffusion model via a reference adapter layer for controlled generation. Extensive experiments on OpenImages-V6 and MSCOCO validate the effectiveness of our CAT-Diffusion.

\subsubsection{Broader Impact.} Recent advancements in generative models (e.g., diffusion models)
have unlocked new realms of creative media generation. However, these innovations also bear the potential to be misused for the generation of deceive content. Our approach may be exploited to inpaint harmful content into images for misinformation spread, and we strongly disapprove of such actions.

\subsubsection{Acknowledgement.} This work was supported by National Natural Science Foundation of China (No. 62172103, 32341012).

%}
%\clearpage  % TODO REVIEW/FINAL: This \clearpage needs to be removed from both review and camera-ready versions.

\bibliographystyle{splncs04}
\bibliography{main_v2}

% \appendix
\begin{figure}[!h]
  \centering
  \includegraphics[width=0.9\linewidth]{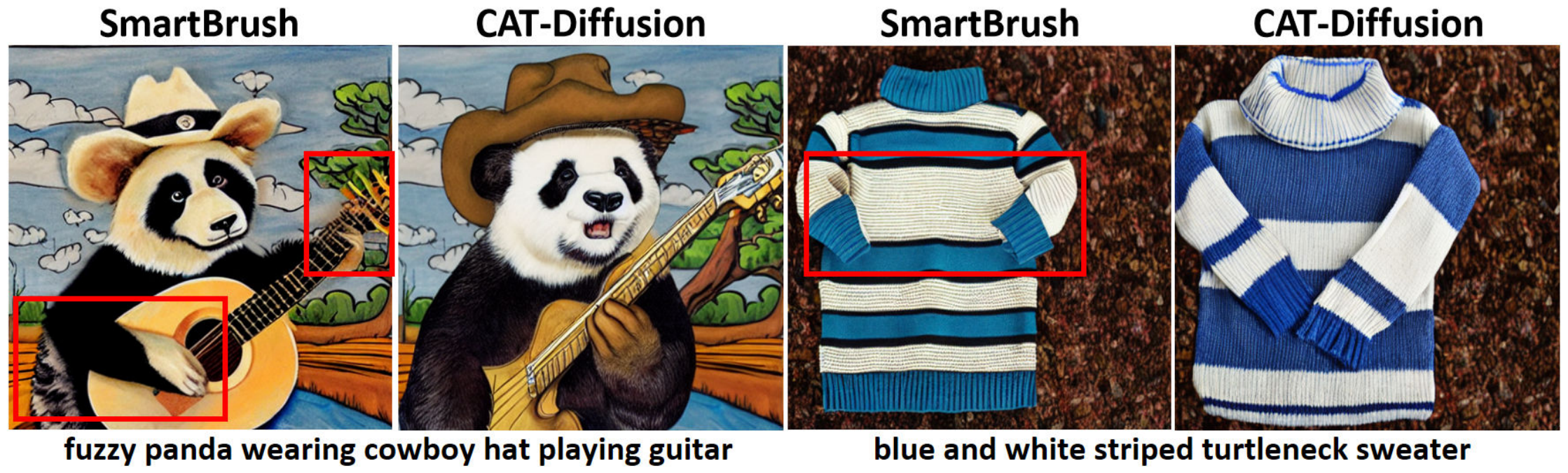}
  \caption{Comparisons with SmartBrush.}
  \label{fig:smartbrush}
\end{figure}

\section{Additional Comparison Results}
\begin{figure}[t]
  \vspace{-0.1in}
  \centering
  \includegraphics[width=1.0\linewidth]{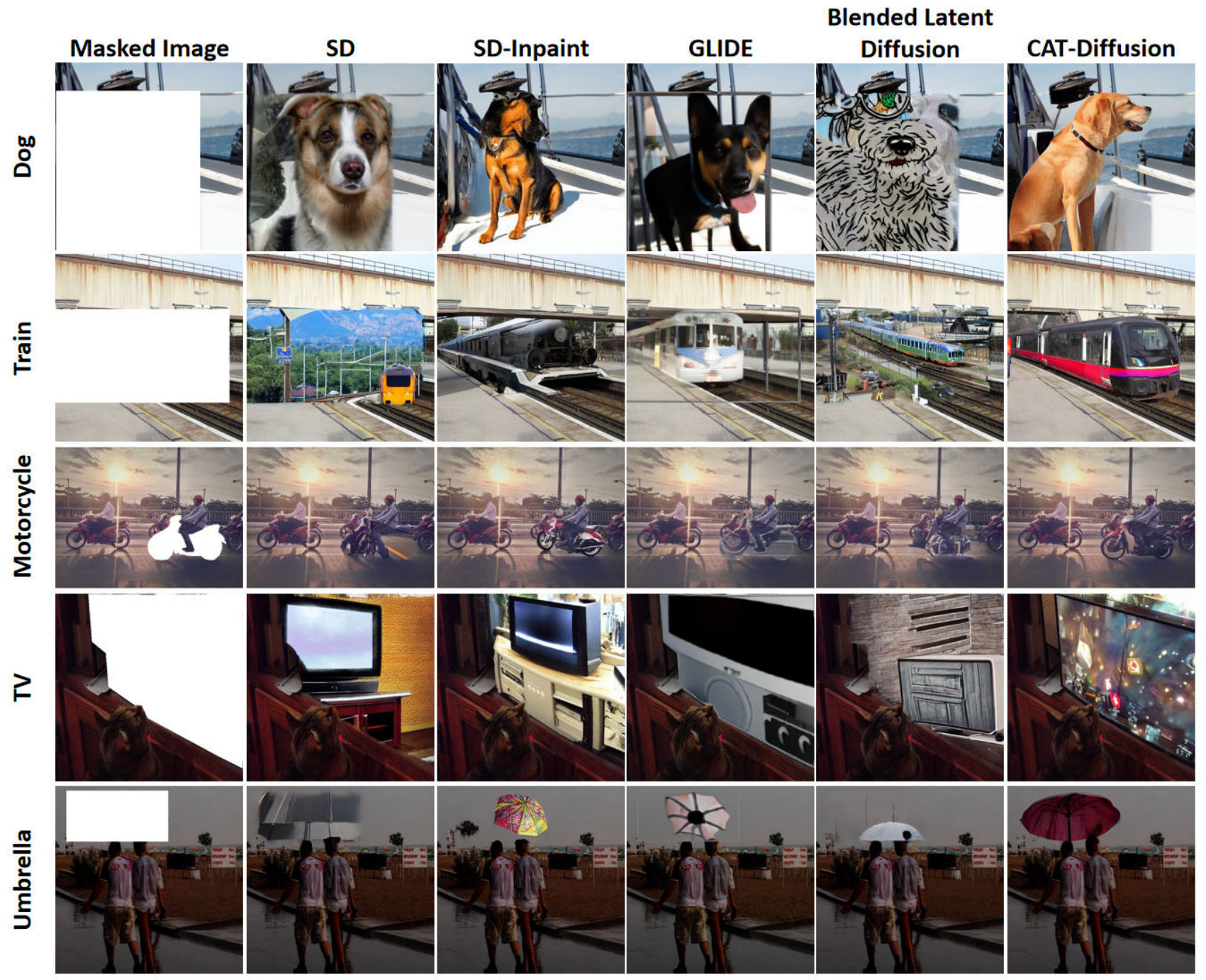}
  \caption{More examples generated by Stable Diffusion (SD), Stable Diffusion Inpainting (SD-Inpaint), GLIDE, Blended Latent Diffusion and our proposed CAT-Diffusion.}
  \label{fig:more_visual}
  \vspace{-0.2in}
\end{figure}

In this section we show additional comparison results including: 1) more comparisons with closed-source SmartBrush in Figure \ref{fig:smartbrush}; 2) more examples generated by different approaches, i.e., Stable Diffusion, Stable Diffusion Inpainting, GLIDE, Blended Latent Diffusion and our proposed CAT-Diffusion in Figure \ref{fig:more_visual}.

\textbf{Comparisons with SmartBrush.}
As shown in Figure \ref{fig:smartbrush}, our CAT-Diffusion depicts better object shapes and visual coherence than SmartBrush. Specifically, the hands of the panda and the upper part of the guitar generated by SmartBrush are defective while the corresponding parts in the image derived from CAT-Diffusion are more satisfactory. Moreover, incorrect geometry is observed in the generated cuffs of the sweater by SmartBrush. Meanwhile, CAT-Diffusion produces more harmonious results.

\textbf{More comparisons with state-of-the-art methods.}
Figure \ref{fig:more_visual} illustrates the inpainted results with both bounding box mask and segmentation mask. It can be easily observed that our proposed CAT-Diffusion generates more pleasing images than the other methods. For example, our CAT-Diffusion accurately fills in a visually coherent dog according to the input prompt on the first row. This again demonstrates the effectiveness of decomposing the typical single-stage object inpainting into two cascaded processes, where the semantic features of the target object are pre-inpainted in a noise-free multi-modal feature space and further leveraged to steer the inpainting diffusion model for high-fieldity object generation across varied noise ratios in the diffusion latent space. 

\end{document}